\documentclass[pmlr,table]{jmlr}

\usepackage{longtable}

\usepackage{booktabs}
\usepackage[load-configurations=version-1]{siunitx} 
\usepackage{float}
\usepackage{todonotes}
\usepackage[super]{nth}

\newcommand{\cs}[1]{\texttt{\char`\\#1}}

\makeatletter
\newcommand\footnoteref[1]{\protected@xdef\@thefnmark{\ref{#1}}\@footnotemark}
\makeatother

\theorembodyfont{\upshape}
\theoremheaderfont{\scshape}
\theorempostheader{:}
\theoremsep{\newline}
\newtheorem*{note}{Note}

\jmlryear{- Under review}
\jmlrworkshop{ NeurIPS 2020 Competition and Demonstration Track}

\title{Learning to run a Power Network Challenge:
a Retrospective Analysis 
}

  \author{\Name{Antoine Marot} 
  \addr {AI Lab, Reseau de Transport d'Electricite (RTE) France }\\
  \Name{Benjamin Donnot}
  \addr {AI Lab, Reseau de Transport d'Electricite (RTE) France }\\
  \Name{Gabriel Dulac-Arnold}
  \addr {Google Research,Brain Team, Google France }\\
  \Name{Adrian Kelly}
  \addr {Electric Power Research Institute (EPRI) Ireland}\\
  \Name{Aidan O'Sullivan}
  \addr {UCL Energy Institute, University College of London}\\
  \Name{Jan Viebahn}
  \addr {Digital and Process Excellence, TenneT}\\
  \Name{Mariette Awad}
  \addr {Electrical \& Computer Engineering Department, American University of Beirut }\\
  \Name{Isabelle Guyon}
  \addr {Chalearn \& University Paris-Saclay France}\\
  \Name{Patrick Panciatici}
  \addr {R\&D Department, Reseau de Transport d'Electricite (RTE) France }\\
  \Name{Camilo Romero}
  \addr {CELEC EP Transelectric Ecuador}
  }

\editor{Hugo Jair Escalante and Katja Hofmann}

\begin{document}
\maketitle
\begin{abstract}

 Power networks, responsible for transporting electricity across large geographical regions, are complex infrastructures on which modern life critically depend. Variations in demand and production profiles, with increasing renewable energy integration, as well as the high voltage network technology, constitute a real challenge for human operators when optimizing electricity transportation while avoiding blackouts. Motivated to  investigate the potential of Artificial Intelligence methods in enabling adaptability in power network operation, we have designed a L2RPN challenge to encourage the development of reinforcement learning solutions to key problems present in the next-generation power networks. The NeurIPS 2020 competition was well received by the international community attracting over 300 participants worldwide.  
 
 The main contribution of this challenge is our proposed comprehensive 'Grid2Op' framework, and associated benchmark, which plays realistic sequential network operations scenarios. The Grid2Op framework, which is open-source and easily re-usable, allows users  to define new environments with its companion GridAlive ecosystem. Grid2Op relies on existing non-linear physical power network simulators and let users create a series of perturbations and challenges that are representative of two important problems: a) the uncertainty resulting from the increased use of unpredictable renewable energy sources, and b) the robustness required with contingent line disconnections. 

In this paper, we give the highlights of the NeurIPS 2020 competition. We present the benchmark suite and analyse the winning solutions, including one super-human performance demonstration. We propose our organizational insights for a successful competition and conclude on open research avenues. Given the challenge success, 
we expect our work will foster research to create more sustainable solutions for power network operations. 
\end{abstract}






\section{Introduction}
\label{sec:intro}

To meet climate goals, power networks are relying more and more on renewable energy generation and decentralising their production.  The traditional tools used by electrical engineers to solve network issues are becoming increasingly inadequate. For further reading on the challenges of future power network operations, readers can refer to the accompanying white paper which introduces the problem, potential solutions  \citep{kelly2020reinforcement}.

There are some promising innovations and recent applications of machine learning (ML) techniques that may offer solutions to these challenges. However the opportunities for ML scientists and researchers to work on the problem was limited by a lack of usable environments, baselines, data, networks and simulators. 

\begin{figure}[htbp]
\centering
\includegraphics[width=\textwidth]{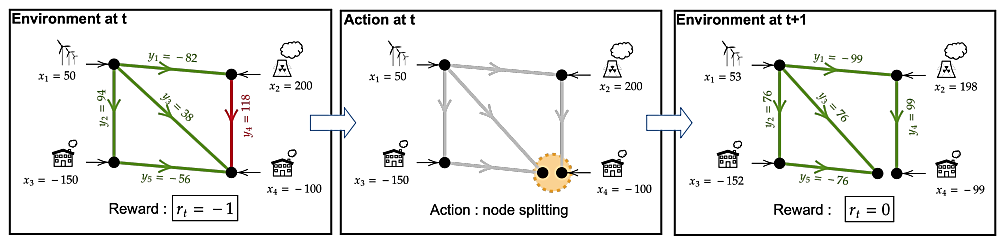}
\caption{\textbf{Power system operation as a complex temporal graph problem:} instead of inferring a given graph properties as in well-known graph problems, we try to find the right graphs over time, among all possible topologies, given the properties that would optimize the flows for safe network operations such as no line overflow.}
\label{fig:L2RPN_RL}
\end{figure}

The ``Learning to run a power network'' (L2RPN) challenge is a series of competitions that model the sequential decision-making environments of real-time power network operations as illustrated in Figure \ref{fig:L2RPN_RL}. It is backed by the open-source GridAlive\footnote{GridAlive ecosystem \url{https://github.com/rte-france/gridAlive}} 
ecosystem and Grid2op\footnote{\label{grid2op}Grid2op package \url{https://github.com/rte-france/Grid2Op}} core framework. 

The aim of L2RPN is to develop a catalyst to foster faster progress in the field, building on advances in AI and reinforcement learning (RL) such as the ImageNet benchmark \citep{deng2009imagenet} for computer vision, and many others in the last decade. The L2RPN challenge aimed at making new large benchmarks for solutions to the real-world problem of complex network operations readily available through the unique Grid2Op framework. The Grid2Op framework has raised awareness of the challenge beyond the power system community, to new, diverse and young communities of scientists and researchers. Additionally, the challenges present an opportunity for the AI community to demonstrate recent breakthroughs which could successfully find applications in other real-world problems, and to eventually go beyond well-studied game environments~\citep{brown2017safe, silver2017mastering, vinyals2019grandmaster, badia2020agent57}. Conversely, the aim is to raise awareness among the power system community about the innovations and potentials of AI and RL algorithms to solve network challenges and to embrace the application of different approaches to traditional problems \citep{prostejovsky2019future}.

Although RL approaches have successfully integrated some real-world applications recently in robot walk \citep{li2021reinforcement}, autonomous vehicle \citep{Cohen} or stratospheric balloon navigation, it has only remained as a research niche in power-systems \citep{glavic2017reinforcement} through years. This is probably because of the lack of both a substantially large benchmark and a widely used framework that we address through L2RPN. Advocates for real-world RL \citep{kidzinski2020artificial, dulacarnold2021challenges, mohanty2020flatland} have indeed made advances in recent years by developing related challenges such as in bio-mechanics and train logistics.  L2RPN also pushes boundaries by framing the challenges over large spatio-temporal decision-making spaces.  Foreseen industrial application could further come as a smart operation recommender system for operators, similar to the one developed for Google data center cooling \citep{Gao}.  

While initial competitions in L2RPN tested the feasibility of developing realistic power network environments \citep{marot2020learning} and the applicability of RL agents \citep{lan2020ai,YHLK2021, subramanian2020exploring}, 
the 2020 L2RPN competition, showcased at NeurIPS 2020, had increased technical complexity. It came with a very large discrete and combinatorial action space dimensions offered by the topological flexibilities of the network. It also introduced a realistically-sized network environment along two tracks, representing two realistic real-life network operation challenges; robustness and adaptability (Section \ref{sec:tasks}). Participation and activity throughout 2020 were steady and there were entries from all over the world. The materials and resources made available to the participants, in particular the Grid2Op framework, are described in Section \ref{sec:resources}. 


Throughout the competition, ML approaches showcased continuous robust and adaptable behaviours over long time horizons. This behaviour was not previously observed by the expert systems, or by optimization methods that are limited by computation time. The results are shown in Section \ref{sec:outcomes} with the best methods described in \ref{sec:top_solutions}. After sharing our organizational insights in  Section \ref{sec:organization}, we conclude on further research avenues to build upon. 


\section{Competition overview}

\label{sec:overview}


This section details the competition design including the problem, tasks and evaluation and the materials that were made available to the participants. Full details of the competition development can be found in \citep{marot2020l2rpn}.

\subsection{Network Operations and Problem Overview}
\label{sec:problem}


The goal of this competition was to control a power network to maintain a supply of electricity to consumers on the network over a given time horizon by avoiding a blackout. In reality, blackouts are caused by cascading failures of overloaded lines which get automatically disconnected. This can cause loss of power to consumers. 
Agents should be designed to optimize the \textbf{cost of  operations} under the safety constraints. The sequential decision making problem is addressed by framing it as a Markov Decision process (MDP) through the lens of RL \citep{SuttonRL}. 

\paragraph{Objectives and game over condition.}
Environments are episodic, running at \textbf{5-minute} resolution over \textbf{a week} (2016 timesteps). The core part of the environment is a power network, with substations (nodes), some containing consumers (load) or production (generation), and interconnecting power lines (edges). The power lines have  different physical characteristics) represented as a \textbf{graph}. The industry standard synthetic \textbf{IEEE 118} network was used for this competition \citep{pena2017extended}.  Realistic production and consumption scenarios were generated for the network. The agents can take actions on the network considering the following constraints:
    
\begin{itemize}
    \item	Events such as maintenance (deterministic) and line disconnections (stochastic and adversarial) can disconnect power lines for several hours. 
    \item	Power lines have intrinsic physical capacities, limiting flows beyond. If a power line is overloaded too long, it automatically disconnects. This can lead to cascading failures.
    \item An agent must wait few hours before reconnecting an incurred line disconnection.
    \item Additionally, to avoid expensive network asset degradation or failure, an agent cannot act too frequently on the same asset or perform too many actions at the same time.
 
\end{itemize}
The Game Over condition is triggered if total demand is not met anymore, that is if some consumption is lost at a substation. More detailed explanation of the physical laws and the physical characteristics of the network can be found in \citep{kelly2020reinforcement}  white paper.
   
\paragraph{Action space.}
There are two allowable actions: cheap topology changes (discrete and combinatorial actions) that allow the agent to modify the graph of the power network and costly production changes from pre-defined generation (continuous actions). Power production can be modified within the physical constraints of each plant. The final action space has more than $70,000$ discrete actions and a $40$-dimensional continuous action space. 
   

\paragraph{Observation space.}
Agents can observe the complete state of the power network at every step. This includes flows on each power line, electricity consumed and produced at each node, power line disconnection duration etc. 
After verification of the previously described constraints, each action is fed into an underlying power flow simulator to run AC powerflow (see \ref{sec:powersim}) to calculate the next network state. Agent also have the opportunity to \textbf{simulate} one's action effect on the current state, to validate their action for instance as operator would do.
But the future remains unknown: anticipating contingencies is not possible, upcoming productions and consumption are stochastic. 

\paragraph{Reward.}
There is a standard score function for the organizers to rank each agent (see \ref{sec:evaluation}). But participants were free to design their own reward function for learning.


\subsection{Physical power flow simulators}
\label{sec:powersim}
For decades, the power system community has dedicated resources to the development of physical simulators and models with defined benchmarks \citep{zimmerman1997matpower}, and there are many on the market.  These simulators calculate power flow for a given network model and a set of initial physical conditions (production and consumption at each node, and topology). Traditionally in the community, ``optimal power flow'' (OPF) problems \citep{babaeinejadsarookolaee2019power} are cast into optimization frameworks which rely on different assumptions, leading to simplifications and tailored simulator implementations. 

By casting the problem as a MDP, no strong assumptions about the network dynamics are necessary anymore. The existing complex simulators which solve the physical equations can be directly leveraged within the Grid2Op framework, letting agents learn through interactions. Aforementioned shortcomings are avoided in L2RPN, the first benchmark that evaluates the continuous time problem while incorporating the complexity. This framework would further apply to upcoming \textbf{cyber-physical} power system environments with multiple agents and automaton interacting on top of the historical physical network layer.

\subsection{Competition design}

\label{sec:design}

\subsubsection{Adaptability and Robustness tracks}
\label{sec:tasks}
\begin{figure}[H]
\includegraphics[width=0.47\linewidth]{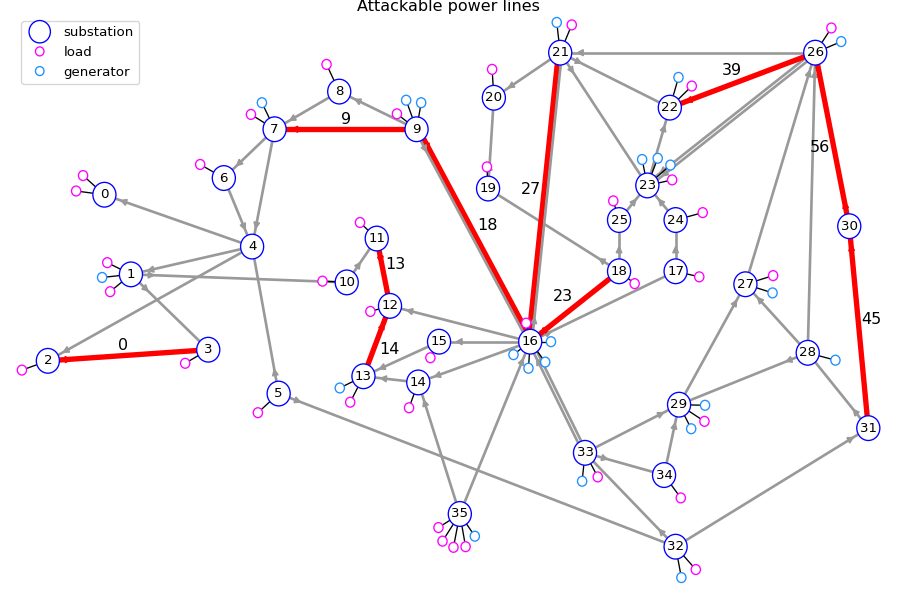}
\includegraphics[width=0.47\linewidth]{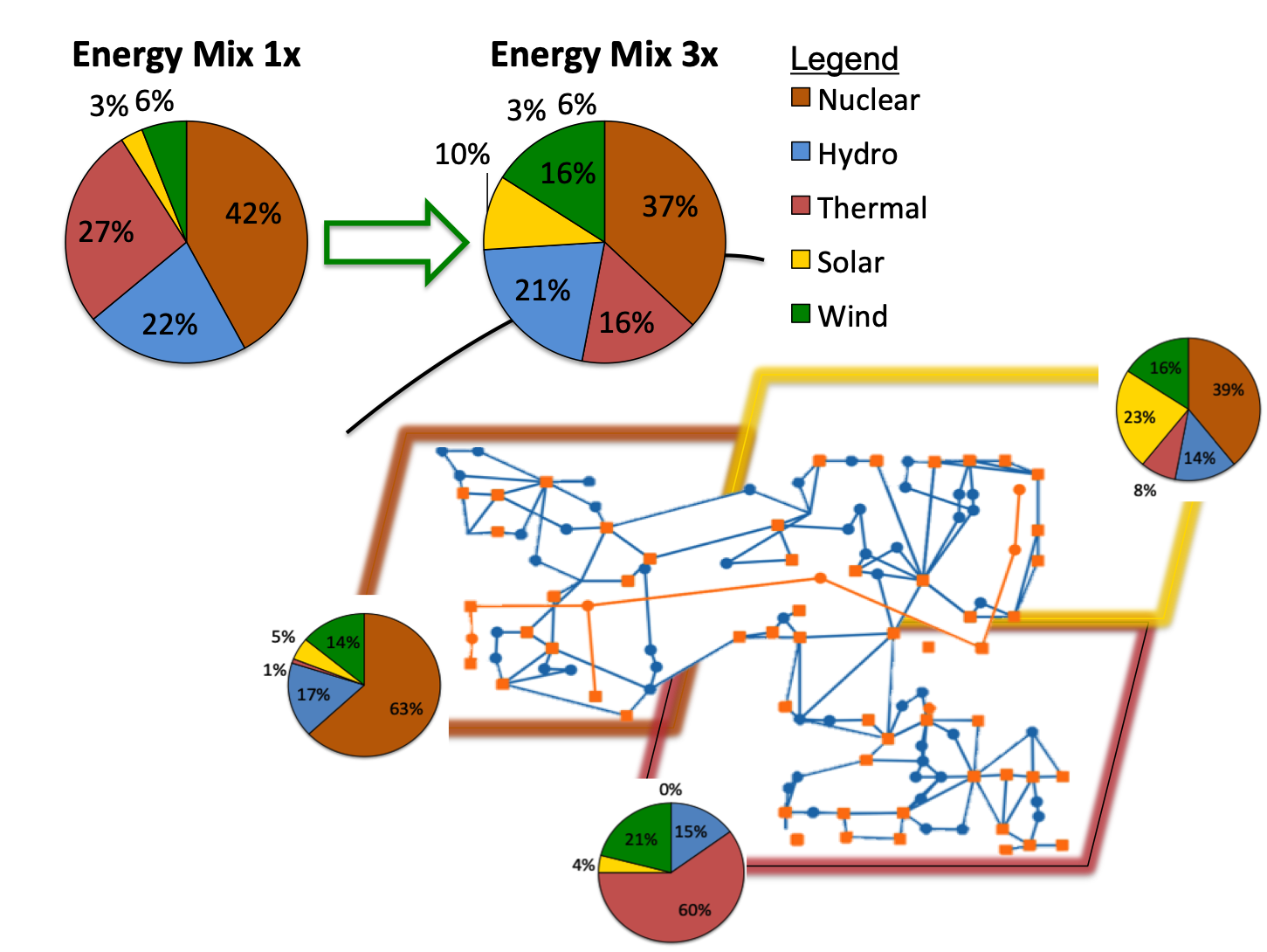}
\caption{Left - Robustness track top right area of 118 bus network, red lines indicate lines under attack. Right - Adaptability track, pie charts are production mixes with varying renewable integration (from 10\% to 30\%) and with differences per area.}

\label{fig_tracks}
\end{figure}

The 2020 competition was divided into two tracks as depicted in Figure \ref{fig_tracks}, which aimed to solve different aspects of the network operation problem. Participants were given hundreds of years of \textit {chronics} which are time-series production, consumption and maintenance schedules to train on. More details can be found in \citep{marot2020l2rpn}.

\paragraph{Robustness Track:}
In this track, agents had to operate the network, maintaining supply to consumers and avoiding overloads while an "opponent" takes adversarial actions. This is achieved through targeted line disconnections, as shown on Figure \ref{fig_tracks} (left), which degrade network operating conditions. The opponent was heuristically designed, disconnecting one of the most loaded lines at random times \citep{omnes2020adversarial}. 

\paragraph{Adaptability Track:}
In this track the ability of agents to cope with different, unseen data distribution within a range of possible distributions was assessed. For training, the participants were given several datasets that were generated from five different energy mix distributions, varying from 10\% (1x) to 30\% (3x) renewables as shown in Figure \ref{fig_tracks}. 
These energy mix distributions were known by participants during training. At test time however, the agents had to respond to unknown energy mix distributions within the 1x-3x range. 

\subsubsection{Evaluation  and ranking}
\label{sec:evaluation}
\paragraph{Three Phases}
The competition was divided into three phases. During the warm-up phase, 
participants could ask any question, some modifications were added to the Grid2op framework based on participants feedback, and new packages could be natively added on the challenge platform. They could already get familiar with the remote submission process, their score appearing over a leaderboard.
The evaluation environment was then fixed. 

The next phase was the feedback phase, which allowed participants to train their model and assess their performance on an unknown validation dataset, getting some granular feedback on their agent survival time per scenario. 
Participants could develop and improve their model iteratively and regularly. 
To avoid over-fitting, the final score was computed during an automatic test phase on a new but similar hidden dataset. 


\paragraph{Evaluation} Participants were evaluated over 24 independent weekly \textbf{test scenarios}, 2 over each month of the year. These undisclosed evaluation datasets were specifically chosen by the competition organizers to be representative of the diverse problems faced by network operators, such as overflow due to high load, or high renewable generation. 

The validation and test datasets were drawn from the same statistical distribution, applying the same methodology to both. The \textbf{score} per scenario was based on the cumulative network operational cost, further normalized as: [-100,0,80,100] = [initial blackout, do-nothing agent score, scenario completed, additional 20\%operational cost optimization].

The leaderboard score was the average score over all 24 scenarios. The running time of the agents on the evaluation platform was eventually limited to $30$ minutes, beyond which -100 score was applied.
More details can be found in \citep{marot2020l2rpn}.

\subsection{Innovative Testbed Framework and Other Competition Resources}
\label{sec:resources}

\paragraph{Grid2Op} L2RPN competitions could use any powersystem simulator to emulate the operational decision making process and cast it as a Markov Decision Process, as explained in \ref{sec:powersim}. The most important innovation and the key tool given to participants was the Grid2op python package\footnoteref{grid2op}. This framework is similar to other recent environments such as RLGC \citep{huang2019adaptive}, pymgrid \citep{henri2020pymgrid}, gym-anm \citep{henry2021gym} or open modelica microgrid \citep{OMG-code2020} in the sense that it allows agents to interact, using the standard "openAI gym" interface \citep{brockman2016openai}, with a power network. 

One major difference for the L2RPN competitions lies in the \textbf{advance action modeling}, 
which utilizes actions on productions and on the topology of the network. By design, Grid2op is also more \textbf{modular}. It allows in particular users to plug in the underlying network simulator model of one's choice, as opposed to other frameworks which depend on one specific simulator. It can in particular get interfaced with \textbf{industrial simulators}, such as the ones in PowSybl framework \footnote{see \url{https://github.com/powsybl}}, enabling faster reuse of developed solutions on more industrial cases. For the L2RPN competitions, the participants could either use the default simulator embedded in the Grid2op platform which is based on pandapower \citep{thurner2018pandapower} or they could use a port in C++ to gain computation speedup.

The Grid2op framework comes equipped with a set of notebooks and extensive technical documentation available online\footnote{see \url{https://grid2op.readthedocs.io/en/latest/}}. This ensured that participants could learn and interact with Grid2op relatively easily, even for newcomers in RL or in power network operation. 
In a post-competition survey (see \ref{sec:survey}), the overall usability of grid2op was rated 3.8 / 5. 
\paragraph{Competition Resources}
Participants could download a "training environment" that they could use as they wished. These environments were made of different generated scenarios, often representing thousands of simulated years of a network (the training environments were different for each track, "Adaptability" and "Robustness" but were identical for each phase: "warm-up", "validation" and "test") as described in \ref{sec:design}.

Participants had to submit code that could be executed on the evaluation server, which is not an easy process. To make it as easy as possible, participants were also given a "\textit{starting kit}" containing descriptions of the environments and the problem to tackle with some examples of valid submissions and a dedicated notebook to help the debugging process. 

\paragraph{L2RPN extra materials}\footnote{Papers, webinar and best team videos can be found at \url{https://l2rpn.chalearn.org/}} To make the competition accessible and attractive to everyone, the problem was thoroughly presented in the "\textit{IEEE Big Data Analytics}"  webinars with live Q\&As, and additional materials were provided.
A white paper was first release on ArXiV \citep{kelly2020reinforcement} to advertise L2RPN to as wide an audience as possible. A "sandbox" competition, on a smaller network from L2RPN 2019 competition, was available as a hands-on. The sandbox was particularly helpful at the beginning, but has been less used ever since.
Finally, we also open sourced a "baselines" repository that contains code of past winning submission (top 1 and 2 of L2RPN 2019, as well as the open sourcing, during the Neurips competition of the winning approach of L2RPN WCCI 2020). More than a dozen of such "toy models" were available to the participants, though only a small half of the participants used this material during the competition, according to the post competition survey.

\section{Outcomes for Participants}
\label{sec:outcomes}
This section presents and explores the results of the competition, from final leaderboards to a detailed analysis of agent behaviour and a description of the winning approaches.

\subsection{Submission Performance Overview}


\begin{table}[t]
  \caption{Top Scores in Robustness (left) and Adaptability tracks (right)}
  \label{leaderboard-robustness}
  \centering
  \begin{minipage}{0.48\textwidth}
      \begin{tabular}{|l|c|r|}
        \hline
        \rowcolor{lightgray} Agent & Score & From  \\
        \hline
        rl\_agent & 59.26 & China \\
        \hline
        binbinchen & 46.89 & China \\
        \hline
        lujixiang & 44.62 & China \\
        \hline
        djmax008 & 43.16 & China/USA\\
        \hline
        tlEEmCES & 41.38 & Germany \\
        \hline
        jianda & 39.68 & Singapore \\
        \hline
        UN\_aiGridOp & 34.72 & Columbia \\
        \hline
        yzm & 33.50 & Singapore \\
        \hline
        Konstantin & 30.38 & Russia \\
        \hline
        panda0246 & 27.36 & China \\
        \hline
        taka & 26.73 & Belgium \\
        \hline
      \end{tabular}
  \end{minipage}
  \hfill
  \begin{minipage}{0.48\textwidth}
      \begin{tabular}{|l|c|r|}
        \hline
        \rowcolor{lightgray} Agent & Score & From  \\
        \hline
        rl\_agent & 51.06 & China \\
        \hline
        kunjietang & 49.32 & China \\
        \hline
        lujixiang & 49.26 & China \\
        \hline
        djmax008 & 42.70 & China/USA\\
        \hline
        TonyS & 28.02 & USA \\
        \hline
        taka & 27.94 & Belgium \\
        \hline
         yzm & 19.76 & Singapore \\
        \hline
        UN\_aiGridOp & 16.76 & Columbia \\
        \hline
        var764 & 15.24 & USA \\
        \hline
        PowerRangers & 12.86 & USA \\
        \hline
        mwasserer & 7.10 & Austria \\
        \hline
      \end{tabular}
  \end{minipage}

\end{table}

By the end of the competition, about 15 participants, using different models, achieved scores above 25 (a very decent base score achieved by our Adaptive Expert System baseline \citep{marot2018expert}) on the Robustness Track, within the 30-min computation time budget. This made the Robustness Track highly competitive. In contrast on the Adaptability Track only about half as many participants surpassed the baseline score. This difference could be explained by a smaller network, hence a smaller observation space dimension, for the Robustness track: it might have required less resources for training. An additional interesting factor was the success of the open-submission phase which increased active participation and performance beyond top-3 teams. As shown on the final leaderboards from Table \ref{leaderboard-robustness}, the 4 top Teams, from Baidu, Huawei, State Grid of China, Geirina, are the same in order on both Robustness and Adaptability Tracks: China definitely made a splash in our competition. While the teams from State Grid of China and Geirina have significant power systems expertise, the performance of Baidu and Huawei was very impressive given their lack of experience in the power sector. 


On the Robustness track, team `RL-agent' from Baidu won by a very comfortable margin. While they reached the 45-point glass ceiling by mid-competition like the two other winning teams, they eventually broke it just 10 days before the end of the competition: they introduced a new successful feature in their approach through evolutionary policy training. 
This is a sign of methodological breakthrough beyond traditional approaches (see \ref{sec:top_teams}). Furthermore, comparing agents to WCCI competition agents, we highlighted that introducing an opponent during training lead to a more robust and eventually better performing agent \citep{omnes2020adversarial} with table shown in Appendix \ref{sec:wcci_table}.

On the adaptability track, scores were a lot tighter. The top scores were achieved earlier in the competition (around October 8 from Figure \ref{fig_timeline}). While the final leader board remained uncertain until the end, no new feature was successfully introduced as a last development to make a strong difference. Even if the scores are already good, there might remain more room for improvement on this adaptability problem, especially under higher variance environments with higher renewable integration, as we will see in Section \ref{sec:results}.


\subsection{Summary of Top 10 Solutions and description for Top 3 methods}
\label{sec:top_solutions}

\begin{table}
  
  \centering

  \begin{tabular}{|l|l|l|c|r|}
    \hline
    \rowcolor{lightgray} Track& Use RL &  Mixed Actions & Graph feat. & Simulate   \\
    \hline
    Robustness & 4/5 (8/10)  & 4/5 (9/10) & 2/5 (5/10) & 5/5 (10/10) \\
    \hline
    Adaptability & 4/5 (9/10)  & 3/5 (5/10) & 1/5 (1/10) & 5/5 (10/10)\\
    \hline
  \end{tabular}

  \caption{Overview of preferred approaches for \textbf{top-5} (resp. \textbf{top-10}) teams.  Mixed actions refers to using both continuous and discrete actions.}
  \label{approch-overview}
\end{table}

Table \ref{approch-overview} summarizes interesting features used by top ranked participants at test time. On both tracks, 4 out of the 5 top teams used RL agent models. Surprisingly, the second best agent of the Adaptability track, kunjietang, used a simple expert system running simulations over an ordered and limited set of 200 appropriate topological actions. On the Robustness track, tlEEmCES team (ranked 5th) did similarly. This sets bounds on the added value of Machine Learning on those problems. Note that all RL-based agents also run simulations online to select an action among the best proposed ones: they did not totally trust RL only decisions.
All teams eventually reduced the action space to a maximum dimension of 1000 unitary actions for exploitation, while dynamically prioritizing them through the learnt model. 
The best teams often used a combination of discrete topological and continuous redispatching actions: redispatching, despite being costly, was most likely necessary on tough scenarios, especially on robustness track (4/5 teams). All teams further used high level expert rules such as \textit{"do-nothing if no lines loaded close to their limits"}, \textit{"reconnect lines in maintenance or attacked when allowed"}, or more generically \textit{"recover initial configuration when possible"}.

As for the observation, they all used line flow usage rate (variables to control) and connectivity status (topological variables to act on). Consumption and production observations were mostly used by agent playing redispatching actions. Calendar variables such as month, day of the week, hour of the day were surprisingly rarely used, except for winning RL\_agent and TonyS. Finally, fewer teams used graph related features of this problem, mostly on track robustness: djMax008, jianda and UN\_aiGridOp used a graph neural network, tlEEmCES used the grid adjacency matrix through an OPF for redispacthing while yzm computed a graph centrality measure feature. Interestingly, Kaist, the winning team from the WCCI competition, did not perform well with their graph neural network approach in this competition, though they did not participate actively.
Considering the reward function, RL\_agent used the simplest possible one: survival duration for the agent. Others used a bit more customized ones, but without much innovations from the ones already available in Grid2op. Reward such as for lujixiang team could be the combination of multiple criteria. But no participants used multi-objective rewards, while available in Grid2Op similarly to safety gym \citep{ray2019benchmarking}.

Overall, while action validation through simulation was a must-have, we believe that discovering robust set of actions during exploration, as well as learning an ability to plan for action sequences as for RL\_agent and binbinchen, made the strongest difference.  

\subsubsection{Top-3 Team approaches}
\label{sec:top_teams}
\paragraph{rl\_agent - \nth{1} place}
This team of experienced 
RL researchers from Baidu won both tracks. They have won similar competitions at Neurips in 2018 and 2019 (Learning To Run and Learning to Move \citep{zhou2019efficient}).  Relying on their "PARL"\footnote{Available at \url{https://github.com/PaddlePaddle/PARL}} framework 
 and the computational resources of Baidu, they were able to train a top performing agent on 500 cores in 30 minutes, displaying seemingly super-human moves\footnote{Video "Advanced Behaviour Analysis of best AI agent" \url{https://www.youtube.com/watch?v=xlqS-CzvMwk}} (see Appendix \ref{sec:scenario_analysis}).

After an action space reduction to 1000 elements with simple expert systems, they initialize a policy parametrized by a feed forward neural network with a few million parameters. 
Then, they train a policy using evolutionary black box optimization \citep{lee2020efficient}. As opposed to most standard deep RL strategies, e.g. Proximal Policy Optimization (PPO) \citep{schulman2017proximal} or Dueling Deep Q Network (DDQN) \citep{wang2016dueling}, they didn't rely exclusively on gradients but on genetic algorithm at last to find the "optimal" weights for their policies, enabling them to eventually overcome large variance issues in reward and value estimation. It improved by 10-points their performance when eventually used in their last-week submitted agents. 


\paragraph{binbinchen - \nth{2} place} The team from Huawei used an RL agent based on an Actor-Critic model with PPO to play topological actions. An expert-guided exploration phase starts by reducing the action space to 200 actions. The resulting expert agent achieved a score around 40-45, tackling the issue of large action space to learn from.  Then they learned a new agent by imitation with a simple feed-forward neural network, performing a bit worse than the expert agent, reaching a score of 40. To  better deal with game constraints requiring planning abilities and achieve long-term rewards, they eventually learn a RL agent, whose actor network is initialized from the imitation agent. This agent achieves a score closer to 50, a 10 point increase from imitation learning reflex agent only.

\paragraph{lujixiang - \nth{3} place} This team used a DDQN-based approach \citep{lan2020ai} that was very similar to the winning team from L2RPN 2019, djmax008.

\subsection{Results details and discussion}
\label{sec:results}
In this section we explore the results achieved to gain insight into the challenges posed by different scenarios and to better understand the behaviour of the top 3 agents. 
\subsubsection{Impact of Different Scenarios on Performance}
\label{sec:scenario_results}

Exploring the variation in the performance of agents across the different test scenarios helps characterise the agents and also the factors that make a scenario more challenging. On some scenarios all agents did equally well, there are other scenarios where only one or two agents performed successfully, and finally some scenarios where all agents failed. 
In the Robustness track, almost all of the 9 unfinished scenarios occurred during winter months, during which the consumption is actually higher leading to a more highly loaded network. In the Adaptability track, the seven unfinished scenarios occur for higher renewable energy mix mix03 (27\% renewables) and mix04 (32\%renewable). Note that mix04 was slightly out of training distribution range (10-30\% renewables), testing for an additional extrapolation capability, which eventually appeared difficult for agents. All scenarios for mix01 (17\% renewables) and mix02 (22\% renewables) are properly managed by at least one agent. Best agents eventually highlight aposteriori the feasibility frontiers, pushing forward our knowledge about the power network flexibilities. Agent episode traces are available for inspection through our Grid2Viz study tool\footnote{Grid2viz tool for best agent study
\url{https://github.com/mjothy/grid2viz}}. More details and survival time heatmaps can also be found in Appendix \ref{sec:result_scenarios}. 


\subsubsection{Agent Behaviour analysis}
The problem posed by this challenge is extremely complex and agents deploy different strategies to respond to the same scenarios. In this section we attempt to gain insight to these strategies by analysing the behaviour of the top 3 agents on Robustness track across three key dimensions, \textbf{action diversity} in the number of substations controlled and number of actions taken and also \textbf{sequence length of the re-configurations} which provides insight into the agent's ability to combine actions in sequences. 

\begin{figure}[t!]
\centerline{\includegraphics[width=0.97\linewidth]{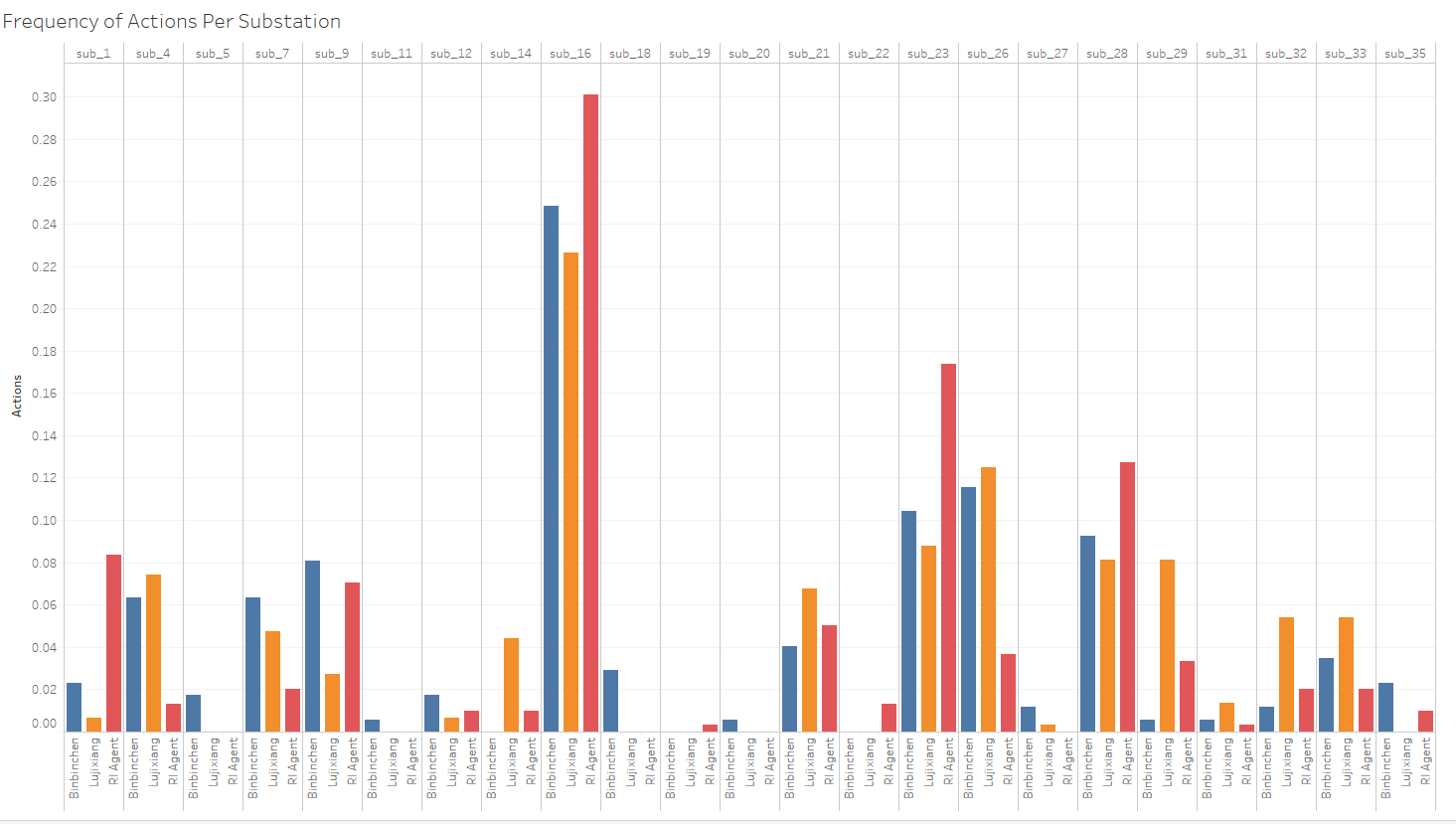}}
\caption{Frequency of unitary substation re-configurations utilized by the top 3 agents across all scenarios.
}
\label{fig_agent_action_diversity}
\end{figure}

\paragraph{Action diversity}
Assessing the agents approach to the substation topology re-configurations reveals distinct patterns of behaviour.
First we note that the agents RLagent and lujixiang utilise significantly more substation re-configurations across all scenarios
(namely 299 and 296, respectively). Agent binbinchen performed about half as many (namely 173). Next, Figure \ref{fig_agent_action_diversity} 
shows how the actions are distributed across the different substations.
While performing the least amount of actions binbinchen (blue) still engages more substations (namely 20) than RLagent (red, 18) and lujixiang (orange, 16), all demonstrating an interesting action diversity.
The substations 16, 23, and 28 represent hubs in the power network and, hence, are often used by all agents. 
Other crucial substations are only relevant to two of the three agents (e.g. substations 4, 26, 29). Finally, both binbinchen and RLagent engage substations that are not used by the other agents (substations 5, 11, 18, 20 and 19, 22, respectively).

\paragraph{Sequence length of reconfigurations}
All agents predominately (for more than $70\%$ of the action sequences, not shown) perform a single substation re-configuration and then do nothing again. Agent binbinchen is especially concise with more than $80\%$ of its action sequences being of length 1. Agent RLagent on the other hand is most pronounced when it comes to executing sequences of length 2-3 (more than $20\%$). Finally, agent lujixiang has a tendency to create long sequences of up to 15 consecutive substation re-configurations (involving up to 7 different substations). The longest sequences of the agents binbinchen and RLagent have a length of 5 and 6, respectively.

In summary, the 3 agents display distinct patterns of Behaviour which demonstrates how the number of controlled substations and the length of action sequences can be balanced in different ways.
Agent binbinchen tends to perform a small amount of actions executed in short sequences but at the same time involves many substations. In contrast, lujixiang operates on a smaller set of substations but performs more actions organised in relatively long sequences. The Behaviour of RLagent is intermediate by engaging a moderate amount of substations in mid-sized sequences, which might be a preferred Behaviour to operators.



\section{Organizational Outcomes}
\label{sec:organization}
\begin{figure}
\centerline{\includegraphics[width=0.97\linewidth]{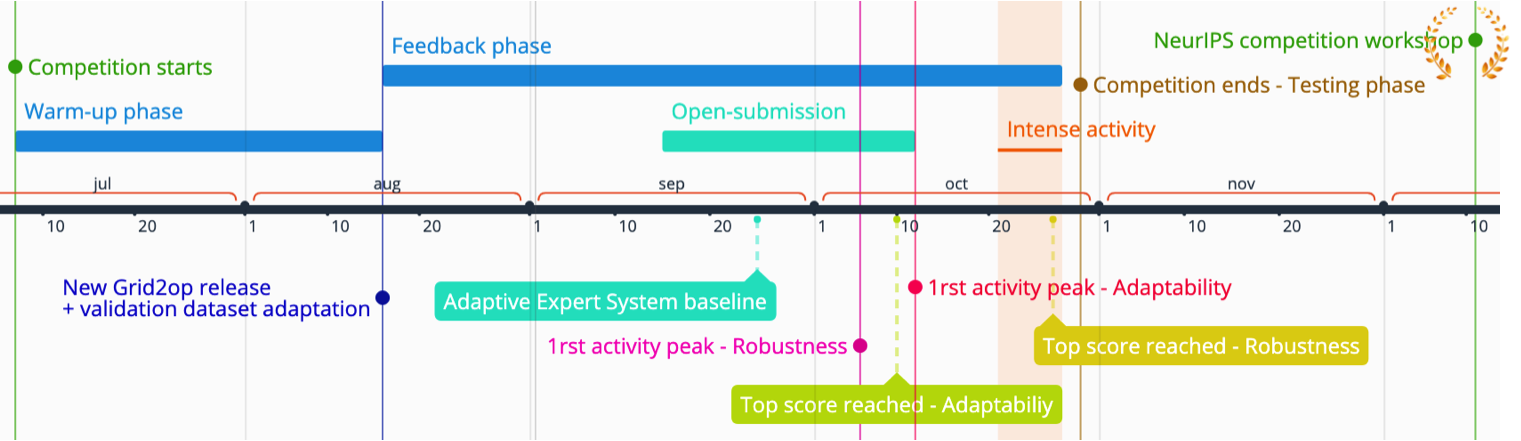}}
\caption{Competition timeline displaying different phases and interesting events}
\label{fig_timeline}
\end{figure}

The competition eventually ran smoothly on Codalab competition platform through our different phases displayed on Figure \ref{fig_timeline}
timeline. We review what contributed to a successful and dynamic competition, while also giving recommendations for possible improvements.

\label{sec:organization}
\subsection{Open-submissions, 2 tracks and 3 phases for a dynamic competition}

\paragraph{Open-submissions:} We aimed at increasing overall participation and interest, by giving everyone a chance to get a prize for their efforts while making their submission public for others. While almost no open-submission happened in the first two weeks, it became successful after we released a new Adaptive Expert System baseline \citep{marot2018expert}), paving the way and increasing participant awareness. Setting an affordable minimum score threshold of 10, 5 best in time and 5 best in score teams won 200\$ and 400\$. 

There was an increase of activity and in the number of new participants by the end of open-submission period. First high activity peaks (over 40 submissions a day) occurred in this period, while we usually see such peaks mainly during the last week of competition. The winning open-submission teams were logically different from the leading teams who did not want to reveal their best agent. The best open-submissions actually combined successfully different parts from other open-submissions, making the competition more and more competitive, bridging some gap with the four leading teams. This also highlights the potential for collaboration in competitions, which could be further explored. 

\paragraph{2 tracks:} The group's previous experience organising competitions  made it possible to streamline the set up process. This allowed two separate tracks to be designed instead of one, on equally important but orthogonal problems for real-world power network, without significant additional investment of resources. It hence let participants chose one were they would be more comfortable or capable with and best compete.
While decoupling the two problems makes it more workable and clearer for participants, 
it also let some team leverage their investment on one (about 2 months), to compete in a shorter time frame with a performant model on the other (only 2 to 3 weeks). Except for the winning teams, teams were mostly different on both tracks. Having two leaderboards, we saw overall an increase of active participation and interest compared to a single track. 
Both tracks ended at the same time in our competition. As peaks of activities occur near the end of the competitions, it might be interesting to actually end a second track two or three weeks after the first one.\vspace{3mm}

Finally, our new warm-up phase was appropriate to get participants engaged, keep improving the framework for competition needs, and adjust validation set difficulty.


\subsection{Effectiveness of Competition Materials}
\label{sec:survey}

We evaluated how useful the competition materials were for the participants, based on what we have learnt through our Discord forum and competition survey.
Among the $64$ unique teams that appeared on the leaderboard and eventually competed, $16$ filled out the survey form made of 35 questions through 4 categories "\textit{competition in general}", "\textit{Grid2op and the competition ecosystem}", "\textit{Codalab, making submissions and competition progression}" and "\textit{Organization and communication during the competition}". 

Overall, all participants seemed to have enjoyed the competition: they rated it 8 / 10 and the Discord forum was greatly appreciated (4.7 / 5) which we recommend to other organizers.  
We indeed think it helps the participants get engaged and get customized help much more quickly: more than $4.700$ messages were posted by both participants and organizers. 
It also allowed us to better understand the needs of the participants and adapt the evaluation servers to their needs with desired software.

As anticipated the submission of code produced some challenges. The most difficult part, for participants, was to provide easily reproducible code that would run on evaluation servers. The question "\textit{How easy was the submission process ?}" was the one with the lowest but yet encouraging average value: 3.6 / 5. Concerning the competition's difficulty,  surprisingly only 1 participant found it difficult among the respondents (only answer below 3)\footnote{Though this statement is probably slightly biased, as probably the people who experienced difficulties did not engage until the end of the competition or were less likely to take the post-competition survey}. Despite this potential bias, we believe this to be representative of the improvement in the submission process, the "getting started" notebooks, and the help provided in the Discord forum. The improvement in the submission process is reflected by the number of submissions made during this competition compared to the previous ones. Indeed, more than 2700 independent submissions were made during the feedback phase, as opposed to approximately $200\sim300$ for the 2019 L2RPN competition and $700\sim800$ for the 2020 WCCI L2RPN competition.


\subsection{Soundness of Evaluation Methodology}
Our scoring over 24 scenarios proved to be robust for the two tracks as the best teams were ranked in same order with similar scores between feedback and test phases, except for one team. The exception was due to overfitting by UN-aiGridOperator Team. Winning the open-submission on the feedback phase, they suffered from overfitting, with a 10-point loss on the test phase. This is one more example of the soundness of having two distinct validation and test sets when evaluating models. The normalized scores per scenarios were more interpretable compared to our first competitions. It helped us better appreciate the performance and Behaviour of agents on each scenarios, giving richer feedback to the participants as well. It also let us compare results on both tracks. 

However there were some opportunities for improvement and lessons learned. A new criteria to filter out hot start scenarios would be useful as analyzed in Appendix \ref{sec:result_scenarios}. In addition, our normalization function around 0 defined by the do-nothing survival time was sometimes a bit too harsh, with agent losing up to 20/100 points in score for only few timesteps. This could be softened to be more representative of the agent performance when it does not do very badly. Finally, 20/100 points per scenario would have been given to agents for continuously optimizing the cost of energy losses, beside surviving. But none won those points, making us wonder how complex this additional task is and what should be the right level of incentives for participants to start tackling it beyond safety. This multi-objective evaluation could be framed differently in the future. 


\section{Conclusion}

The L2RPN NeurIPS competition has demonstrated how encouraging collaboration between the AI and power systems communities can lead to the advancement of novel and innovative AI methods for one of the most significant and challenging problems in the power sector, grid operation. The winning team RlAgent was a team of researchers from Baidu, a tech company, with no domain expertise in power systems. This highlights the value in making tools like the Grid2Op platform available, which lower the barriers for entry in this domain, and the benefit of running global competitions, which attract the interest of leading AI researchers. As well as being a problem with valuable real-world applications the L2RPN challenge is also pushing the state of the art of AI forward as a uniquely challenging problem which involves graph topology, complex physics and high dimensional action spaces which have non-local effects throughout the network. Model-based approaches such as physics-informed ones could be relevant in that regard to make further progress. 

Yet, the behaviour demonstrated  by the best agents was very encouraging with interesting successful long action sequences even under strong perturbations, showcasing some desired robustness. It showed that it could augment knowledge of grid flexibility beyond human capability, which would already be a valuable application for operators when applied on their grid.   It needs to be highlighted that in this problem setting identifying the `optimal' response that an agent should take is almost impossible due to the combinatiorial nature of the problem. Deeper analysis of agent behaviour will have revealing insights for power systems operators. Solutions could be extended to capture further complexity; currently no continuous power loss optimization was performed by agents other than satisfying safety considerations: this would be a highly impactful next addition. As the power system evolves to meet the decarbonisation targets it will become increasingly complex for operators who will require new tools and support in their decision making. An additional research avenue is trustworthy human-machine interactions as we envision AI assistants for operators \citep{marot2020towards} rather than a fully autonomous system: it comes as the next research direction showcased in our new "L2RPN with Trust" competition\footnote{"L2RPN with Trust"
\url{https://icaps21.icaps-conference.org/Competitions/}}.

It is hoped that this challenge opens up further fruitful and important collaborations between the power system community, which has a host of complex decision making problems which need solutions, and the AI community who can benefit from the opportunity to apply Reinforcement and Deep Learning methods to real-world problems. Through the growing open-source GridAlive ecosystem, researchers are encouraged to share new agent baselines to be benchmarked and studied, as well as to build new relevant environments. It is hoped that new competitions will develop given their attractive catalyst effect for research, and that the insights discussed in this paper will prove useful to new organizers. Lastly, L2RPN interactive materials are a valuable resource  for education and teaching.

\section{Acknowledgements}
We thank our sponsors RTE, EPRI, Google Research, IQT Labs and UCL Energy Institute for financing the cumulative 30 000\$ prizes over both tracks. We also thank Google Cloud Platform for their compute credits, Chalearn for using their Codalab compeition platform as well as Florent Carli for making possible to run the competition. We finally thank Loïc Omnes and Jean Grizet for taking part in the development of the competition, as well as Yang Weng, Kishan Guddanti and Zigfrid Hampel  for promoting the competition. 

\bibliographystyle{plain} 
\bibliography{references}

\section{Appendix}
\subsection{Comparison of WCCI and NeurIPS competition agents}
\label{sec:wcci_table}
WCCI and NeurIPS competitions differ only by the introduction of an opponent.
This table \ref{tab:track1_result_comparision} is taken from our paper \citep{omnes2020adversarial} and highlights differences in performance when the opponent was available during training or not. Not only are agents stronger on the NeurIPS Robustness competition when they were given the opponent for training, they also outperform the WCCI agent on the WCCI competition. 
Plus, while the scores for the NeurIPS agents are only a little bit lower on the NeurIPS competition, which is expected since that competition uses an opponent and is thus harder to deal with, the performance of WCCI agent zenghsh3 (actually from Baidu as well) drops much more dramatically when confronted to the opponent.

\begin{table}[H]
  \caption{Agents results on the WCCI and NeurIPS competitions}
  \label{agents-results}
  \centering
  \begin{tabular}{llll}
    \toprule
    \multicolumn{3}{c}{}{Results}                   \\
    \cmidrule{2-4}
    Agent     & From competition & WCCI   & NeurIPS \\
    \midrule
    rl\_agnet         & NeurIPS  & 71.21  & 61.05     \\
    binbinchen        & NeurIPS  & 63.97  & 52.42     \\
    lujixiang         & NeurIPS  & 73.73  & 45.00     \\
    zenghsh3          & WCCI     & 58.21  & 19.10     \\
    reco\_powerline   & Baseline & 25.75  & 10.76     \\
    do\_nothing       & Baseline & 0.00   & 0.00      \\
    \bottomrule
  \end{tabular}
  \label{tab:track1_result_comparision}
\end{table}

\subsection{Detailed results over scenarios}
\label{sec:result_scenarios}
\begin{figure}
\includegraphics[width=8cm]{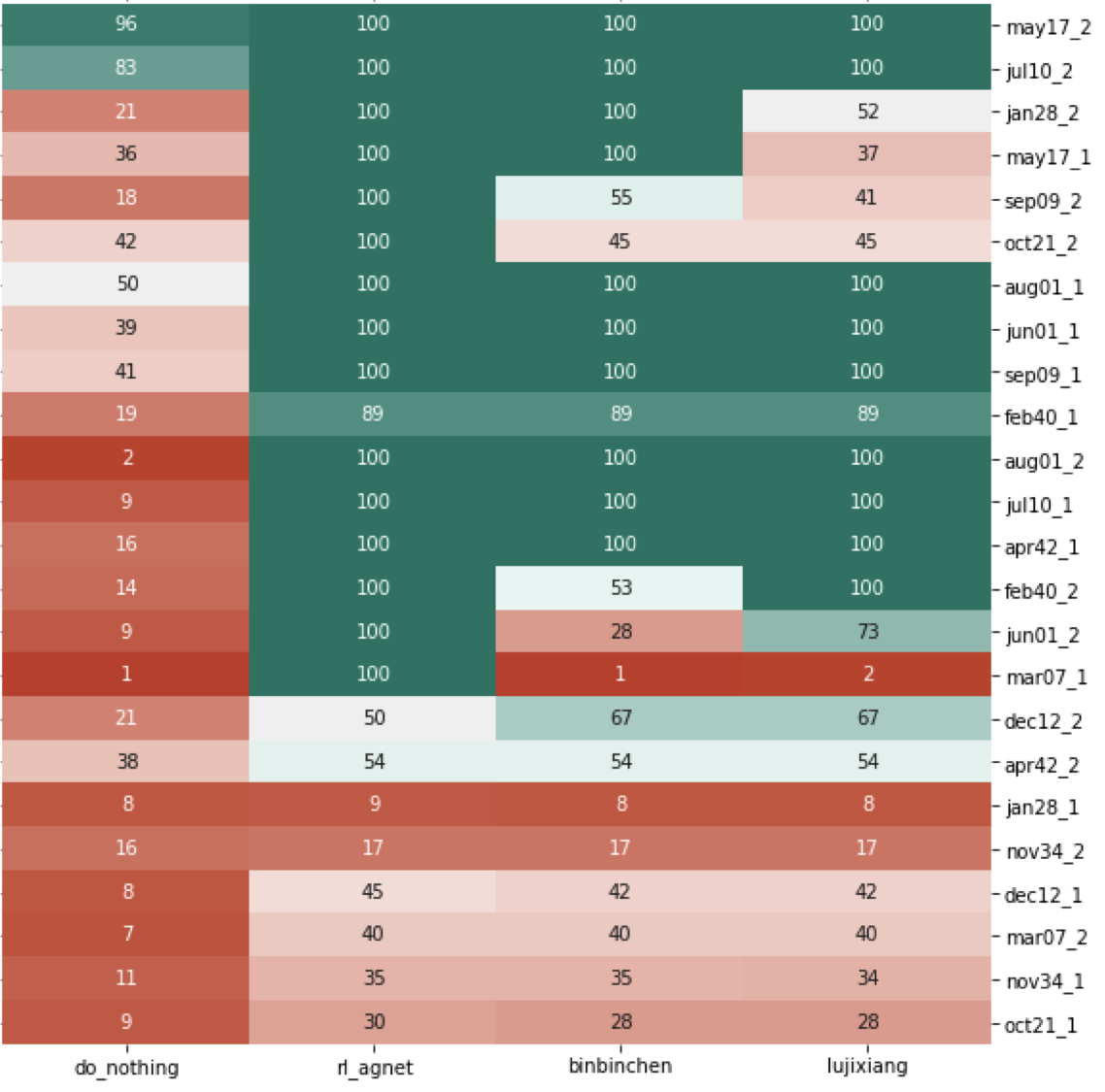}
\includegraphics[width=8cm]{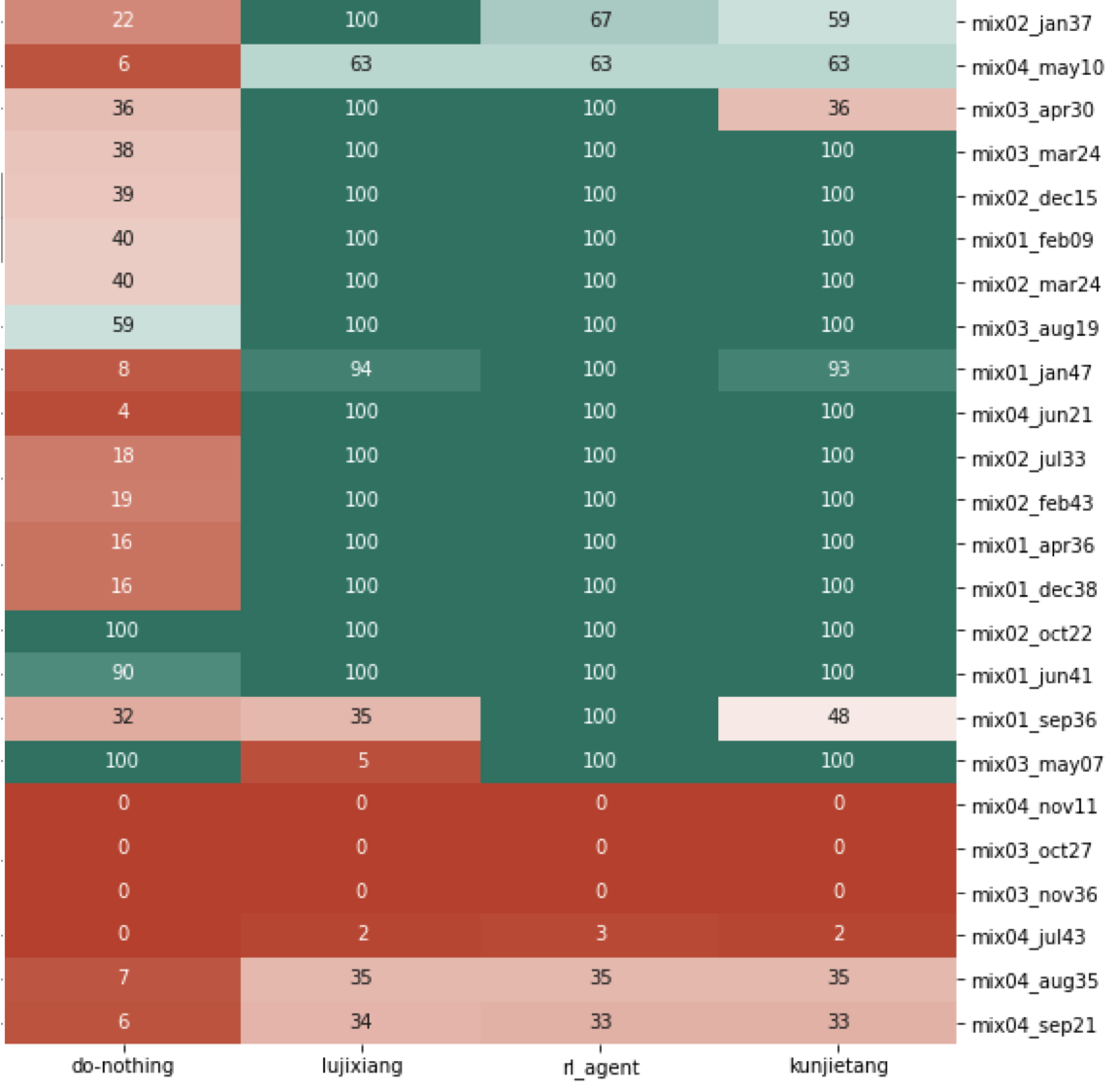}
\caption{Survival time percentage heatmaps (red for early blackout and green for complete episode) for do-nothing agent baseline and top 3 agent, over the 24 weekly test scenarios for robustness (left) and adaptability (right) tracks }
\label{fig_heatmaps}
\end{figure}

We discuss the feasibility of unfinished scenarios looking at the survival time heatmaps of Figure \ref{fig_heatmaps}. For robustness track, we find that a set of scenarios (Jan 28.1 + nov34.1 + apr 42.2 + oct 21.1) looked tough but feasible at the time of gameover using only topological actions, another set (Feb40.1 + March 07.2 + dec12.2) would require combined redispatching type of action, while a last set (Nov34.2 + dec 12.1) looked eventually nearly impossible without new flexibilities (voltage control, load shedding) not available during the competition. 
For Adaptability track, three scenarios (mix04\_nov11, mix03\_oct27 and mix03\_nov36) actually appeared infeasible because of a harsh start with many already existing overloads to solve without possible anticipation. Other mix04 scenarios appear quite tough but possible, all presenting a high renewable production evacuation issue at substation 61, sometimes combined to simultaneous overloads in the Southern part of the network.
Without any ground-truth, it is often hard apriori to know for some cases if they will indeed be feasible as we have seen here, but this can be refined aposteriori.

\subsection{Hard Scenario analysis on Robustness track}
\label{sec:scenario_analysis}
\begin{figure}
\includegraphics[width=8cm]{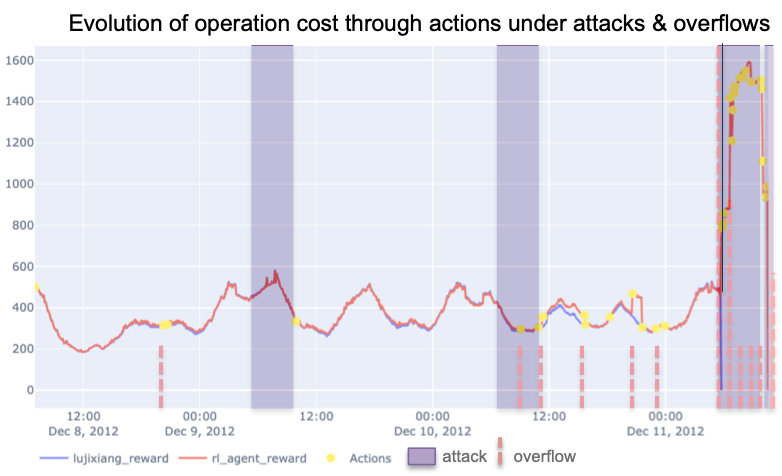}
\includegraphics[width=8cm]{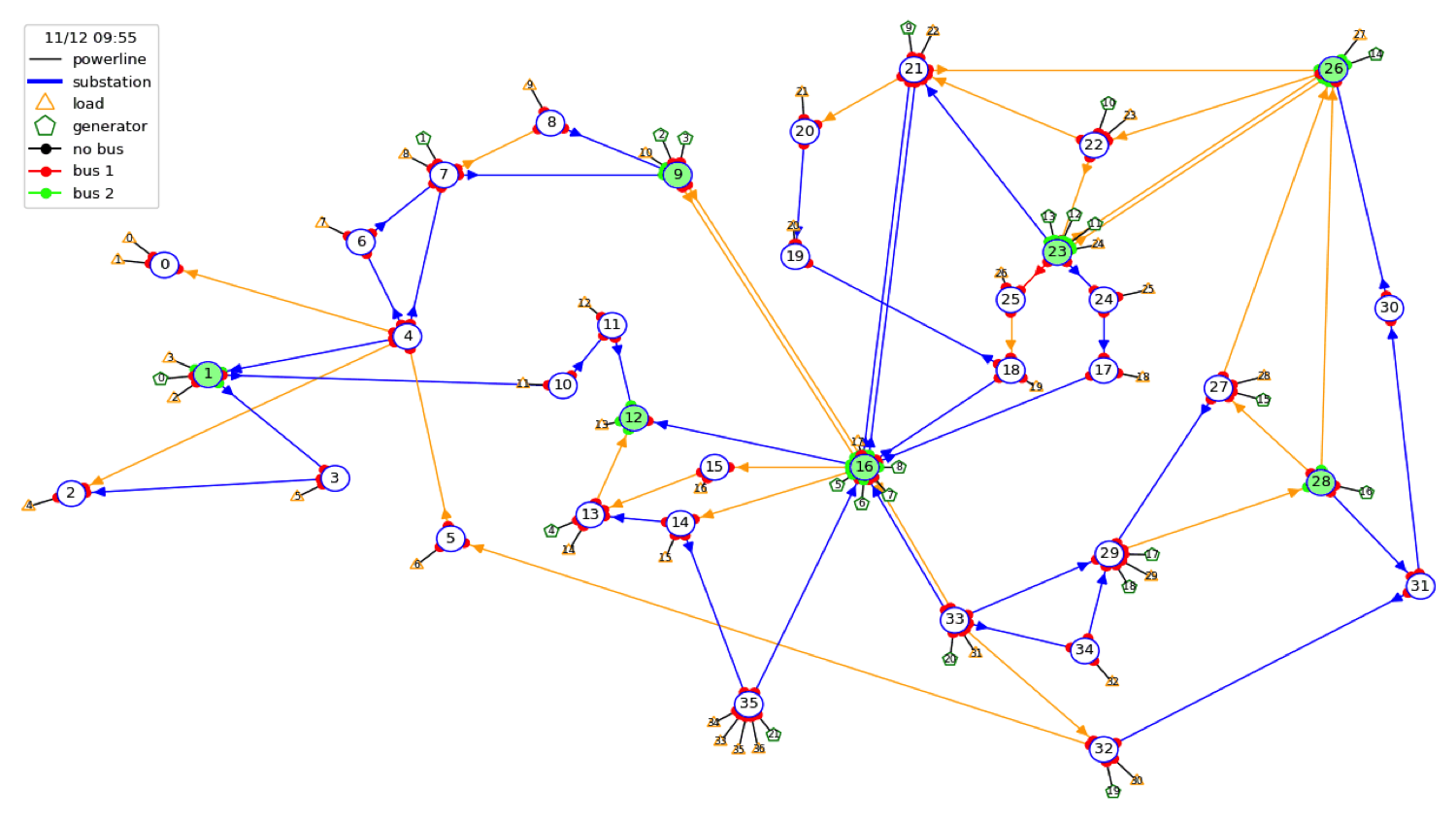}
\caption{Network operation cost over time (left) for best agents RLagent and lujixiang in scenario dec12.1, overcoming highlighted periods of attack and overflows. Only RLagent manages the third and hard attack by playing very effectively complex sequence of actions, sometimes in a superhuman fashion, over 7 substations (highlighted on the right). It nevertheless fails after a new attack a bit later.}
\label{fig_agent_Behaviour}
\end{figure}

Scenarios where the best agent survive longer than other agents, but still fail later, are interesting ones to study from a Behaviour perspective. We here discuss our analysis on such a scenario dec12.1. RL\_agent is eventually the only one to survive on December 11th over a strong attack at 6am against a very high voltage line 26-30, as shown on Figure \ref{fig_agent_Behaviour}. This line was supplying a fair amount of consumption located at the top right of the network. This large powerflow is therefore redipatched over smaller capacity lines connected to this area. Overloads appear on some of these lines, without much margin left on others: the flows needs to be balance timely and precisely, making the task quite complex. Over this 4-hour short period of time, RL\_agent execute lots of actions (20), through sometimes impressive complex 3 to 5 combined action sequences. It successfully adapts to new overloads appearing because of a continuously increasing consumption up to 10am.
As combined-action depth superior to 3 is hard to study for human operators, some sequences showcased by the agent can be today considered \textbf{super-human}. 

Ahead of the competition, we anticipated that "it remains largely unknown today how much flexibility an existing network topology can eventually offer when controlled widely at a high frequency because of its non-linear combinatorial complexity". This result shows that AI could indeed help operators augment their knowledge and use of network topology flexibility in the future. Conversely, RL\_agent eventually fails over a new attack that a human could most probably have solved. AI can hence be better regarded as an \textbf{assistant} for the operator here rather than a fully autonomous agent in charge of the network.

\iffalse
\section{Introduction}
\label{sec:intro}

This is a sample article that uses the \textsf{jmlr} class with
the \texttt{wcp} class option.  Please follow the guidelines in
this sample document as it can help to reduce complications when
combining the articles into a book. Please avoid using obsolete
commands, such as \verb|\rm|, and obsolete packages, such as
\textsf{epsfig}.\footnote{See
\url{http://www.ctan.org/pkg/l2tabu}}

Please also ensure that your document will compile with PDF\LaTeX.
If you have an error message that's puzzling you, first check for it
at the UK TUG FAQ
\url{https://texfaq.org/FAQ-man-latex}.  If
that doesn't help, create a minimal working example (see
\url{https://www.dickimaw-books.com/latex/minexample}) and post
to somewhere like TeX on StackExchange
(\url{https://tex.stackexchange.com/}) or the LaTeX Community Forum
(\url{https://latex.org/forum/}).

\begin{note}
This is an numbered theorem-like environment that was defined in
this document's preamble.
\end{note}

\subsection{Sub-sections}

Sub-sections are produced using \verb|\subsection|.

\subsubsection{Sub-sub-sections}

Sub-sub-sections are produced using \verb|\subsubsection|.

\paragraph{Sub-sub-sub-sections}

Sub-sub-sub-sections are produced using \verb|\paragraph|.
These are unnumbered with a running head.

\subparagraph{Sub-sub-sub-sub-sections}

Sub-sub-sub-sub-sections are produced using \verb|\subparagraph|.
These are unnumbered with a running head.

\section{Cross-Referencing}

Always use \verb|\label| and \verb|\ref| (or one of the commands
described below) when cross-referencing.  For example, the next
section is Section~\ref{sec:math}. The \textsf{jmlr} class
provides some convenient cross-referencing commands:
\verb|\sectionref|, \verb|\equationref|, \verb|\tableref|,
\verb|\figureref|, \verb|\algorithmref|, \verb|\theoremref|,
\verb|\lemmaref|, \verb|\remarkref|, \verb|\corollaryref|,
\verb|\definitionref|, \verb|\conjectureref|, \verb|\axiomref|,
\verb|\exampleref| and \verb|\appendixref|. The argument of these
commands may either be a single label or a comma-separated list
of labels. Examples:

Referencing sections: \sectionref{sec:math} or
\sectionref{sec:intro,sec:math} or
\sectionref{sec:intro,sec:math,sec:tables,sec:figures}.

Referencing equations: \equationref{eq:trigrule} or
\equationref{eq:trigrule,eq:df} or
\equationref{eq:trigrule,eq:f,eq:df,eq:y}.

Referencing tables: \tableref{tab:operatornames} or
\tableref{tab:operatornames,tab:example} or
\tableref{tab:operatornames,tab:example,tab:example-booktabs}.

Referencing figures: \figureref{fig:nodes} or
\figureref{fig:nodes,fig:teximage} or
\figureref{fig:nodes,fig:teximage,fig:subfigex} or
\figureref{fig:circle,fig:square}.

Referencing algorithms: \algorithmref{alg:gauss} or
\algorithmref{alg:gauss,alg:moore} or
\algorithmref{alg:gauss,alg:moore,alg:net}.

Referencing theorem-like environments: \theoremref{thm:eigenpow},
\lemmaref{lem:sample}, \remarkref{rem:sample}, 
\corollaryref{cor:sample}, \definitionref{def:sample},
\conjectureref{con:sample}, \axiomref{ax:sample} and
\exampleref{ex:sample}.

Referencing appendices: \appendixref{apd:first} or
\appendixref{apd:first,apd:second}.

\section{Equations}
\label{sec:math}

The \textsf{jmlr} class loads the \textsf{amsmath} package, so
you can use any of the commands and environments defined there.
(See the \textsf{amsmath} documentation for further
details.\footnote{Either \texttt{texdoc amsmath} or
\url{http://www.ctan.org/pkg/amsmath}})

Unnumbered single-lined equations should be displayed using
\verb|\[| and \verb|\]|. For example:
\[E = m c^2\]
Numbered single-line equations should be displayed using the
\texttt{equation} environment. For example:
\begin{equation}\label{eq:trigrule}
\cos^2\theta + \sin^2\theta \equiv 1
\end{equation}
This can be referenced using \verb|\label| and \verb|\equationref|.
For example, \equationref{eq:trigrule}.

Multi-lined numbered equations should be displayed using the
\texttt{align} environment.\footnote{For reasons why you 
shouldn't use the obsolete \texttt{eqnarray} environment, see
Lars Madsen, \emph{Avoid eqnarray!} TUGboat 33(1):21--25, 2012.} For example:
\begin{align}
f(x) &= x^2 + x\label{eq:f}\\
f'(x) &= 2x + 1\label{eq:df}
\end{align}
Unnumbered multi-lined equations should be displayed using the
\texttt{align*} environment. For example:
\begin{align*}
f(x) &= (x+1)(x-1)\\
&= x^2 - 1
\end{align*}
If you want to mix numbered with unnumbered lines use the
align environment and suppress unwanted line numbers with
\verb|\nonumber|. For example:
\begin{align}
y &= x^2 + 3x - 2x + 1\nonumber\\
&= x^2 + x + 1\label{eq:y}
\end{align}
An equation that is too long to fit on a single line can be
displayed using the \texttt{split} environment. 
Text can be embedded in an equation using \verb|\text| or
\verb|\intertext| (as used in \theoremref{thm:eigenpow}).
See the \textsf{amsmath} documentation for further details.

\subsection{Operator Names}
\label{sec:op}

Predefined operator names are listed in \tableref{tab:operatornames}.
For additional operators, either use \verb|\operatorname|,
for example $\operatorname{var}(X)$ or declare it with
\verb|\DeclareMathOperator|, for example
\begin{verbatim}
\DeclareMathOperator{\var}{var}
\end{verbatim}
and then use this new command. If you want limits that go above and
below the operator (like \verb|\sum|) use the starred versions
(\verb|\operatorname*| or \verb|\DeclareMathOperator*|).

\begin{table}[htbp]
\floatconts
  {tab:operatornames}%
  {\caption{Predefined Operator Names (taken from 
   \textsf{amsmath} documentation)}}%
  {%
\begin{tabular}{rlrlrlrl}
\cs{arccos} & $\arccos$ &  \cs{deg} & $\deg$ &  \cs{lg} & $\lg$ &  \cs{projlim} & $\projlim$ \\
\cs{arcsin} & $\arcsin$ &  \cs{det} & $\det$ &  \cs{lim} & $\lim$ &  \cs{sec} & $\sec$ \\
\cs{arctan} & $\arctan$ &  \cs{dim} & $\dim$ &  \cs{liminf} & $\liminf$ &  \cs{sin} & $\sin$ \\
\cs{arg} & $\arg$ &  \cs{exp} & $\exp$ &  \cs{limsup} & $\limsup$ &  \cs{sinh} & $\sinh$ \\
\cs{cos} & $\cos$ &  \cs{gcd} & $\gcd$ &  \cs{ln} & $\ln$ &  \cs{sup} & $\sup$ \\
\cs{cosh} & $\cosh$ &  \cs{hom} & $\hom$ &  \cs{log} & $\log$ &  \cs{tan} & $\tan$ \\
\cs{cot} & $\cot$ &  \cs{inf} & $\inf$ &  \cs{max} & $\max$ &  \cs{tanh} & $\tanh$ \\
\cs{coth} & $\coth$ &  \cs{injlim} & $\injlim$ &  \cs{min} & $\min$ \\
\cs{csc} & $\csc$ &  \cs{ker} & $\ker$ &  \cs{Pr} & $\Pr$
\end{tabular}\par
\begin{tabular}{rlrl}
\cs{varlimsup} & $\varlimsup$ 
& \cs{varinjlim} & $\varinjlim$\\
\cs{varliminf} & $\varliminf$ 
& \cs{varprojlim} & $\varprojlim$
\end{tabular}
}
\end{table}

\section{Vectors and Sets}
\label{sec:vec}

Vectors should be typeset using \cs{vec}. For example $\vec{x}$.
The \textsf{jmlr} class also provides \cs{set} to typeset a
set. For example $\set{S}$.

\section{Floats}
\label{sec:floats}

Floats, such as figures, tables and algorithms, are moving
objects and are supposed to float to the nearest convenient
location. Please don't force them to go in a particular place. In
general it's best to use the \texttt{htbp} specifier and don't
put the figure or table in the middle of a paragraph (that is
make sure there's a paragraph break above and below the float).
Floats are supposed to have a little extra space above and below
them to make them stand out from the rest of the text. This extra
spacing is put in automatically and shouldn't need modifying.

To ensure consistency, please \emph{don't} try changing the format of the caption by doing
something like:
\begin{verbatim}
\caption{\textit{A Sample Caption.}}
\end{verbatim}
or
\begin{verbatim}
\caption{\em A Sample Caption.}
\end{verbatim}
You can, of course, change the font for individual words or 
phrases, for example:
\begin{verbatim}
\caption{A Sample Caption With Some \emph{Emphasized Words}.}
\end{verbatim}

\subsection{Tables}
\label{sec:tables}

Tables should go in the \texttt{table} environment. Within this
environment use \verb|\floatconts| (defined by \textsf{jmlr})
to set the caption correctly and center the table contents.

\begin{table}[htbp]
\floatconts
  {tab:example}%
  {\caption{An Example Table}}%
  {\begin{tabular}{ll}
  \bfseries Dataset & \bfseries Result\\
  Data1 & 0.12345\\
  Data2 & 0.67890\\
  Data3 & 0.54321\\
  Data4 & 0.09876
  \end{tabular}}
\end{table}

If you want horizontal rules you can use the \textsf{booktabs}
package which provides the commands \verb|\toprule|, 
\verb|\midrule| and \verb|\bottomrule|. For example, see
\tableref{tab:example-booktabs}.

\begin{table}[hbtp]
\floatconts
  {tab:example-booktabs}
  {\caption{A Table With Horizontal Lines}}
  {\begin{tabular}{ll}
  \toprule
  \bfseries Dataset & \bfseries Result\\
  \midrule
  Data1 & 0.12345\\
  Data2 & 0.67890\\
  Data3 & 0.54321\\
  Data4 & 0.09876\\
  \bottomrule
  \end{tabular}}
\end{table}

If you want vertical lines as well, you can't use the
\textsf{booktabs} commands as there'll be some unwanted gaps.
Instead you can use \LaTeX's \verb|\hline|, but the rows may
appear a bit cramped.  You can add extra space above or below a
row using \verb|\abovestrut| and \verb|\belowstrut|. For example,
see \tableref{tab:example-hline}.

\begin{table}[htbp]
\floatconts
  {tab:example-hline}
  {\caption{A Table With Horizontal and Vertical Lines}}%
  {%
    \begin{tabular}{|l|l|}
    \hline
    \abovestrut{2.2ex}\bfseries Dataset & \bfseries Result\\\hline
    \abovestrut{2.2ex}Data1 & 0.12345\\
    Data2 & 0.67890\\
    Data3 & 0.54321\\
    \belowstrut{0.2ex}Data4 & 0.09876\\\hline
    \end{tabular}
  }
\end{table}

If you want to align numbers on their decimal point, you can
use the \textsf{siunitx} package. For example, see
\tableref{tab:example-siunitx}. For further details see the
\textsf{siunitx} documentation\footnote{Either \texttt{texdoc
siunitx} or \url{http://www.ctan.org/pkg/siunitx}}.

\begin{table}[htbp]
\floatconts
  {tab:example-siunitx}
  {\caption{A Table With Numbers Aligned on the Decimal Point}}
  {\begin{tabular}{lS[tabformat=3.5]}
  \bfseries Dataset & {\bfseries Result}\\
  Data1 & 0.12345\\
  Data2 & 10.6789\\
  Data3 & 50.543\\
  Data4 & 200.09876
  \end{tabular}}
\end{table}

If the table is too wide, you can adjust the inter-column
spacing by changing the value of \verb|\tabcolsep|. For
example:
\begin{verbatim}
\setlength{\tabcolsep}{3pt}
\end{verbatim}
If the table is very wide but not very long, you can use the
\texttt{sidewaystable} environment defined in the
\textsf{rotating} package (so use \verb|\usepackage{rotating}|).
If the table is too long to fit on a page, you should use the
\texttt{longtable} environment defined in the \textsf{longtable}
package (so use \verb|\usepackage{longtable}|).

\subsection{Figures}
\label{sec:figures}

Figures should go in the \texttt{figure} environment. Within this
environment, use \verb|\floatconts| to correctly position the
caption and center the image. Use \verb|\includegraphics|
for external graphics files but omit the file extension. Do not
use \verb|\epsfig| or \verb|\psfig|. If you want to scale the
image, it's better to use a fraction of the line width rather
than an explicit length. For example, see \figureref{fig:nodes}.

\begin{figure}[htbp]
\floatconts
  {fig:nodes}
  {\caption{Example Image}}
  {\includegraphics[width=0.5\linewidth]{images/nodes}}
\end{figure}

If your image is made up of \LaTeX\ code (for example, commands
provided by the \textsf{pgf} package) you can include it using
\cs{includeteximage} (defined by the \textsf{jmlr} class). This
can be scaled and rotated in the same way as \cs{includegraphics}.
For example, see \figureref{fig:teximage}.

\begin{figure}[htbp]
\floatconts
  {fig:teximage}
  {\caption{Image Created Using \LaTeX\ Code}}
  {\includeteximage[angle=45]{images/teximage}}
\end{figure}

If the figure is too wide to fit on the page, you can use the
\texttt{sidewaysfigure} environment defined in the
\textsf{rotating} package.

Don't use \verb|\graphicspath|. If the images are contained in
a subdirectory, specify this when you include the image, for
example \verb|\includegraphics{figures/mypic}|.

\subsubsection{Sub-Figures}
\label{sec:subfigures}

Sub-figures can be created using \verb|\subfigure|, which is
defined by the \textsf{jmlr} class. The optional argument allows
you to provide a subcaption. The label should be placed in the
mandatory argument of \verb|\subfigure|. You can reference the
entire figure, for example \figureref{fig:subfigex}, or you can
reference part of the figure using \verb|\figureref|, for example
\figureref{fig:circle}. Alternatively you can reference the
subfigure using \verb|\subfigref|, for example
\subfigref{fig:circle,fig:square} in \figureref{fig:subfigex}.

\begin{figure}[htbp]
\floatconts
  {fig:subfigex}
  {\caption{An Example With Sub-Figures.}}
  {%
    \subfigure[A Circle]{\label{fig:circle}%
      \includegraphics[width=0.2\linewidth]{images/circle}}%
    \qquad
    \subfigure[A Square]{\label{fig:square}%
      \includegraphics[width=0.2\linewidth]{images/square}}
  }
\end{figure}

By default, the sub-figures are aligned on the baseline.
This can be changed using the second optional argument
of \verb|\subfigure|. This may be \texttt{t} (top), \texttt{c}
(centered) or \texttt{b} (bottom). For example, the subfigures
\subfigref{fig:circle2,fig:square2} in \figureref{fig:subfigex2}
both have \verb|[c]| as the second optional argument.

\begin{figure}[htbp]
\floatconts
  {fig:subfigex2}
  {\caption{Another Example With Sub-Figures.}}
  {%
    \subfigure[A Small Circle][c]{\label{fig:circle2}%
      \includegraphics[width=0.1\linewidth]{images/circle}}%
    \qquad
    \subfigure[A Square][c]{\label{fig:square2}%
      \includegraphics[width=0.2\linewidth]{images/square}}
  }
\end{figure}

\subsection{Sub-Tables}
\label{sec:subtables}
There is an analogous command \verb|\subtable| for sub-tables.
It has the same syntax as \verb|\subfigure| described above.
You can reference the table using \verb|\tableref|, for example
\tableref{tab:subtabex} or you can reference part of the table,
for example \tableref{tab:ab}. Alternatively you can reference the
subtable using \verb|\subtabref|, for example
\subtabref{tab:ab,tab:cd} in \tableref{tab:subtabex}.

\begin{table}[htbp]
\floatconts
 {tab:subtabex}
 {\caption{An Example With Sub-Tables}}
 {%
   \subtable{%
     \label{tab:ab}%
     \begin{tabular}{cc}
     \bfseries A & \bfseries B\\
     1 & 2
     \end{tabular}
   }\qquad
   \subtable{%
     \label{tab:cd}%
     \begin{tabular}{cc}
     \bfseries C & \bfseries D\\
     3 & 4\\
     5 & 6
     \end{tabular}
   }
 }
\end{table}

By default, the sub-tables are aligned on the top.
This can be changed using the second optional argument
of \verb|\subtable|. This may be \texttt{t} (top), \texttt{c}
(centered) or \texttt{b} (bottom). For example, the sub-tables
\subtabref{tab:ab2,tab:cd2} in \tableref{tab:subtabex2}
both have \verb|[c]| as the second optional argument.

\begin{table}[htbp]
\floatconts
 {tab:subtabex2}
 {\caption{Another Example With Sub-Tables}}
 {%
   \subtable[][c]{%
     \label{tab:ab2}%
     \begin{tabular}{cc}
     \bfseries A & \bfseries B\\
     1 & 2
     \end{tabular}
   }\qquad
   \subtable[][c]{%
     \label{tab:cd2}%
     \begin{tabular}{cc}
     \bfseries C & \bfseries D\\
     3 & 4\\
     5 & 6
     \end{tabular}
   }
 }
\end{table}

\subsection{Algorithms}
\label{sec:algorithms}

Enumerated textual algorithms can be displayed using the
\texttt{algorithm} environment. Within this environment, use
use an \texttt{enumerate} or nested \texttt{enumerate} environments.
For example, see \algorithmref{alg:gauss}. Note that algorithms
float like figures and tables.

\begin{algorithm}[htbp]
\floatconts
{alg:gauss}
{\caption{The Gauss-Seidel Algorithm}}
{
\begin{enumerate}
  \item For $k=1$ to maximum number of iterations
    \begin{enumerate}
      \item For $i=1$ to $n$
        \begin{enumerate}
        \item $x_i^{(k)} = 
          \frac{b_i - \sum_{j=1}^{i-1}a_{ij}x_j^{(k)}
          - \sum_{j=i+1}^{n}a_{ij}x_j^{(k-1)}}{a_{ii}}$
        \item If $\|\vec{x}^{(k)}-\vec{x}^{(k-1)} < \epsilon\|$,
          where $\epsilon$ is a specified stopping criteria, stop.
      \end{enumerate}
    \end{enumerate}
\end{enumerate}
}
\end{algorithm}

You can use \verb|\caption| and \verb|\label| without using
\verb|\floatconts| (as in \algorithmref{alg:moore}).

If you'd rather have the same numbering throughout the algorithm
but still want the convenient indentation of nested 
\texttt{enumerate} environments, you can use the
\texttt{enumerate*} environment provided by the \textsf{jmlr}
class. For example, see \algorithmref{alg:moore}.

\begin{algorithm}
\caption{Moore's Shortest Path}\label{alg:moore}
Given a connected graph $G$, where the length of each edge is 1:
\begin{enumerate*}
  \item Set the label of vertex $s$ to 0
  \item Set $i=0$
  \begin{enumerate*}
    \item \label{step:locate}Locate all unlabelled vertices 
          adjacent to a vertex labelled $i$ and label them $i+1$
    \item If vertex $t$ has been labelled,
    \begin{enumerate*}
      \item[] the shortest path can be found by backtracking, and 
      the length is given by the label of $t$.
    \end{enumerate*}
    otherwise
    \begin{enumerate*}
      \item[] increment $i$ and return to step~\ref{step:locate}
    \end{enumerate*}
  \end{enumerate*}
\end{enumerate*}
\end{algorithm}

Pseudo code can be displayed using the \texttt{algorithm2e}
environment. This is defined by the \textsf{algorithm2e} package
(which is automatically loaded) so check the \textsf{algorithm2e}
documentation for further details.\footnote{Either \texttt{texdoc
algorithm2e} or \url{http://www.ctan.org/pkg/algorithm2e}}
For an example, see \algorithmref{alg:net}.

\begin{algorithm2e}
\caption{Computing Net Activation}
\label{alg:net}
\KwIn{$x_1, \ldots, x_n, w_1, \ldots, w_n$}
\KwOut{$y$, the net activation}
$y\leftarrow 0$\;
\For{$i\leftarrow 1$ \KwTo $n$}{
  $y \leftarrow y + w_i*x_i$\;
}
\end{algorithm2e}

\section{Description Lists}

The \textsf{jmlr} class also provides a description-like 
environment called \texttt{altdescription}. This has an
argument that should be the widest label in the list. Compare:
\begin{description}
\item[add] A method that adds two variables.
\item[differentiate] A method that differentiates a function.
\end{description}
with
\begin{altdescription}{differentiate}
\item[add] A method that adds two variables.
\item[differentiate] A method that differentiates a function.
\end{altdescription}

\section{Theorems, Lemmas etc}
\label{sec:theorems}

The following theorem-like environments are predefined by
the \textsf{jmlr} class: \texttt{theorem}, \texttt{example},
\texttt{lemma}, \texttt{proposition}, \texttt{remark}, 
\texttt{corollary}, \texttt{definition}, \texttt{conjecture}
and \texttt{axiom}. You can use the \texttt{proof} environment
to display the proof if need be, as in \theoremref{thm:eigenpow}.

\begin{theorem}[Eigenvalue Powers]\label{thm:eigenpow}
If $\lambda$ is an eigenvalue of $\vec{B}$ with eigenvector
$\vec{\xi}$, then $\lambda^n$ is an eigenvalue of $\vec{B}^n$
with eigenvector $\vec{\xi}$.
\begin{proof}
Let $\lambda$ be an eigenvalue of $\vec{B}$ with eigenvector
$\xi$, then
\begin{align*}
\vec{B}\vec{\xi} &= \lambda\vec{\xi}
\intertext{premultiply by $\vec{B}$:}
\vec{B}\vec{B}\vec{\xi} &= \vec{B}\lambda\vec{\xi}\\
\Rightarrow \vec{B}^2\vec{\xi} &= \lambda\vec{B}\vec{\xi}\\
&= \lambda\lambda\vec{\xi}\qquad
\text{since }\vec{B}\vec{\xi}=\lambda\vec{\xi}\\
&= \lambda^2\vec{\xi}
\end{align*}
Therefore true for $n=2$. Now assume true for $n=k$:
\begin{align*}
\vec{B}^k\vec{\xi} &= \lambda^k\vec{\xi}
\intertext{premultiply by $\vec{B}$:}
\vec{B}\vec{B}^k\vec{\xi} &= \vec{B}\lambda^k\vec{\xi}\\
\Rightarrow \vec{B}^{k+1}\vec{\xi} &= \lambda^k\vec{B}\vec{\xi}\\
&= \lambda^k\lambda\vec{\xi}\qquad
\text{since }\vec{B}\vec{\xi}=\lambda\vec{\xi}\\
&= \lambda^{k+1}\vec{\xi}
\end{align*}
Therefore true for $n=k+1$. Therefore, by induction, true for all
$n$.
\end{proof}
\end{theorem}

\begin{lemma}[A Sample Lemma]\label{lem:sample}
This is a lemma.
\end{lemma}

\begin{remark}[A Sample Remark]\label{rem:sample}
This is a remark.
\end{remark}

\begin{corollary}[A Sample Corollary]\label{cor:sample}
This is a corollary.
\end{corollary}

\begin{definition}[A Sample Definition]\label{def:sample}
This is a definition.
\end{definition}

\begin{conjecture}[A Sample Conjecture]\label{con:sample}
This is a conjecture.
\end{conjecture}

\begin{axiom}[A Sample Axiom]\label{ax:sample}
This is an axiom.
\end{axiom}

\begin{example}[An Example]\label{ex:sample}
This is an example.
\end{example}

\section{Color vs Grayscale}
\label{sec:color}

It's helpful if authors supply grayscale versions of their
images in the event that the article is to be incorporated into
a black and white printed book. With external PDF, PNG or JPG
graphic files, you just need to supply a grayscale version of the
file. For example, if the file is called \texttt{myimage.png},
then the gray version should be \texttt{myimage-gray.png} or
\texttt{myimage-gray.pdf} or \texttt{myimage-gray.jpg}. You don't
need to modify your code. The \textsf{jmlr} class checks for
the existence of the grayscale version if it is print mode 
(provided you have used \verb|\includegraphics| and haven't
specified the file extension).

You can use \verb|\ifprint| to determine which mode you are in.
For example, in \figureref{fig:nodes}, the 
\ifprint{dark gray}{purple} ellipse represents an input and the
\ifprint{light gray}{yellow} ellipse represents an output.
Another example: {\ifprint{\bfseries}{\color{red}}important text!}

You can use the class option \texttt{gray} to see how the
document will appear in gray scale mode. \textcolor{blue}{Colored
text} will automatically be converted to gray scale.

The \textsf{jmlr} class loads the \textsf{xcolor}
package, so you can also define your own colors. For example:
\ifprint
  {\definecolor{myred}{gray}{0.5}}%
  {\definecolor{myred}{rgb}{0.5,0,0}}%
\textcolor{myred}{XYZ}.

The \textsf{xcolor} class is loaded with the \texttt{x11names}
option, so you can use any of the x11 predefined colors (listed
in the \textsf{xcolor} documentation\footnote{either 
\texttt{texdoc xcolor} or \url{http://www.ctan.org/pkg/xcolor}}).

\section{Citations and Bibliography}
\label{sec:cite}

The \textsf{jmlr} class automatically loads \textsf{natbib}.
This sample file has the citations defined in the accompanying
BibTeX file \texttt{jmlr-sample.bib}. For a parenthetical
citation use \verb|\citepp|. For example
\citepp{guyon-elisseeff-03}. For a textual citation use
\verb|\citept|. For example \citept{guyon2007causalreport}.
Both commands may take a comma-separated list, for example
\citept{guyon-elisseeff-03,guyon2007causalreport}.

These commands have optional arguments and have a starred
version. See the \textsf{natbib} documentation for further
details.\footnote{Either \texttt{texdoc natbib} or
\url{http://www.ctan.org/pkg/natbib}}

The bibliography is displayed using \verb|\bibliography|.

\acks{Acknowledgements go here.}

\bibliography{jmlr-sample}

\appendix

\section{First Appendix}\label{apd:first}

This is the first appendix.

\section{Second Appendix}\label{apd:second}

This is the second appendix.
\fi

\end{document}